\documentclass{article}

\usepackage{microtype}
\usepackage{graphicx}
\usepackage{booktabs} 
\usepackage{subcaption} 
\usepackage{stfloats} 

\usepackage{hyperref}

\usepackage[accepted]{template/icml2020}

\usepackage{ifthen}
\newcommand\hide[1]{}
\newboolean{show_todos}
\newboolean{show_notes}
\setboolean{show_todos}{true}
\setboolean{show_notes}{true}
\newcommand\todo[1]{\ifthenelse{\boolean{show_todos}}{\textcolor{red}{\textbf{ToDo: }#1}}{\hide{#1}}}
\newcommand\note[1]{\ifthenelse{\boolean{show_notes}}{\textcolor{blue}{#1}}{\hide{#1}}}
\newcommand\laura[1]{\ifthenelse{\boolean{show_notes}}{\textcolor{orange}{\textbf{Laura:} #1}}{\hide{#1}}}
\newcommand\tim[1]{\ifthenelse{\boolean{show_notes}}{\textcolor{orange}{\textbf{Tim:} #1}}{\hide{#1}}}

\begin{document}

\twocolumn[
\icmltitle{Towards Map-Based Validation of Semantic Segmentation Masks}




\begin{icmlauthorlist}
\icmlauthor{Laura von Rueden}{f}
\icmlauthor{Tim Wirtz}{f}
\icmlauthor{Fabian Hueger}{vw}
\icmlauthor{Jan David Schneider}{vw}
\icmlauthor{Christian Bauckhage}{f}
\end{icmlauthorlist}

\icmlaffiliation{f}{Fraunhofer Center for Machine Learning, Fraunhofer IAIS, Sankt Augustin, Germany}
\icmlaffiliation{vw}{Volkswagen Group Automation, Wolfsburg, Germany}

\icmlcorrespondingauthor{Laura von Rueden}{laura.von.rueden@iais.fraunhofer.de}

\icmlkeywords{Machine Learning, Semantic Segmentation, Street Map, Validation, ICML, Autonomous Driving}

\vskip 0.3in
]



\printAffiliationsAndNotice{}  

\begin{abstract}
Artificial intelligence for autonomous driving must meet strict requirements on safety and robustness.
We propose to validate machine learning models for self-driving vehicles not only with given ground truth labels, but also with additional a-priori knowledge.
In particular, we suggest to validate the drivable area in semantic segmentation masks using given street map data.
We present first results, which indicate that prediction errors can be uncovered by map-based validation.


\end{abstract}

\section{Introduction}

Environmental perception is important for autonomous vehicles in order to assess the surrounding traffic scene and understand its context~\cite{campbell2010autonomous, pendleton2017perception}.
A key component is semantic segmentation,
which assigns pixel-wise pre-defined class labels to the input images from vehicle's cameras.
Current algorithms use machine and deep learning techniques to build models that predict semantic segments and surpass classic computer vision techniques in terms of performance~\cite{garcia2018survey, feng2019deep}.

The development of artificial intelligence systems brings certain challenges, especially when they are applied in safety-critical areas.
Building deep neural networks that generalize well and are robust often comes with the need for large amounts of ground truth data, which is typically acquired in expensive manual labelling processes.
To ensure the safety of AI-based systems, mechanisms that support a trustworthy development like interpretability, auditing and risk assessment are discussed with growing interest~\cite{brundage2020toward}.

The validation of machine learning models is particularly important in the area of highly automated driving, for example the identification and mitigation of risks of potential functional insufficiencies in neural networks used for perception~\cite{burton2017making}.
Since the perception component is responsible for the first assessment of the vehicle's surroundings, the detection and reduction of errors in this component can increase the reliability of the resulting environment model.
Proposed approaches for mitigation are the detection of prediction uncertainties~\cite{kendall2017uncertainties} and the estimation of an according error propagation~\cite{mcallister2017concrete}.
Although state-of-the-art neural networks for semantic segmentation achieve promising results, it can still be observed that certain objects of the drivable area are not detected correctly. Moreover, smaller networks being used for embedded purposes are often comparatively less accurate than state-of-the-art networks with arbitrary size. As an example, roads and pedestrian walks could be mixed up in difficult lighting conditions or unusual terrain.

\begin{figure}[t]
\centering
\includegraphics[width=\columnwidth]{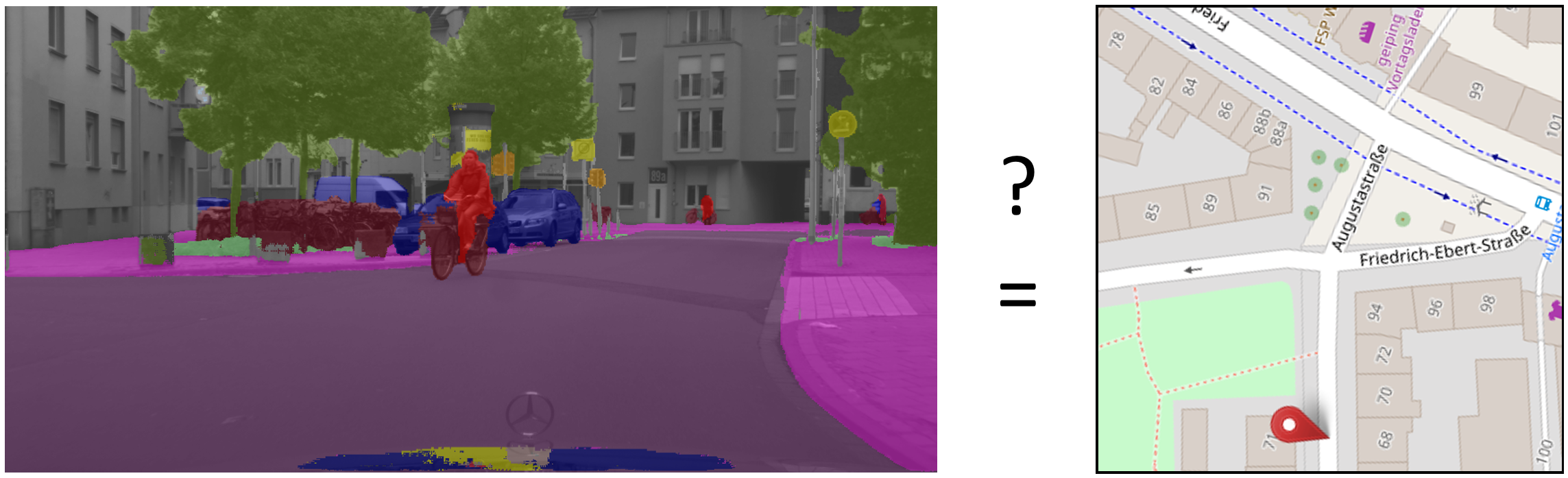}
\caption{
\textbf{Research question.} Can predicted semantic segmentation masks be validated with a-priori knowledge from street maps? The left image shows a segmentation of a traffic scene in the Cityscapes dataset~\cite{cordts2016cityscapes} and the right image shows the corresponding map~\cite{OpenStreetMap}. Here, an intersection to the right, which is shown in the map, is not reflected in the segmentation.
}
\label{fig:intro}
\end{figure}

We propose to support the goal of safe artificial intelligence in autonomous driving by applying the idea of informed machine learning~\cite{vonrueden2020informed} and validate learned models with a-priori knowledge.
In this paper, we suggest to compare semantic segmentation masks to the structured semantic information in street maps, as illustrated in Figure~\ref{fig:intro}, and present a novel method that computes the overlap of drivable area between the segmentation output and the map.
Our approach is inspired by how human drivers would perceive environments: When they find themselves in a new environment, they often consult external knowledge sources such as street maps and compare what they see in their vicinity to what they see on the map.

Related work comprises approaches for multi-modal perception for autonomous driving, combining the inputs from various driving data~\cite{feng2019deep}.
The combination of camera inputs and street maps for semantic segmentation has already been used to assign geographical addresses to detect buildings~\cite{ardeshir2015geo}, or to build conditional random fields for scene understanding~\cite{wang2015holistic, diaz2016lifting}.
The integration of general geographic or geometric a-priori knowledge in perception tasks has been investigated in versatile forms, for example as shape priors for object localization~\cite{murthy2017shape}, temporal priors for revisited locations~\cite{schroeder2019using}, or spatial relation graphs for object detection~\cite{xu2019spatial}.
However, to the best of our knowledge, there is not yet an approach that uses street maps for a validation of neural network based semantic segmentation masks.

\begin{figure*}[t]
\centering
\includegraphics[width=\textwidth]{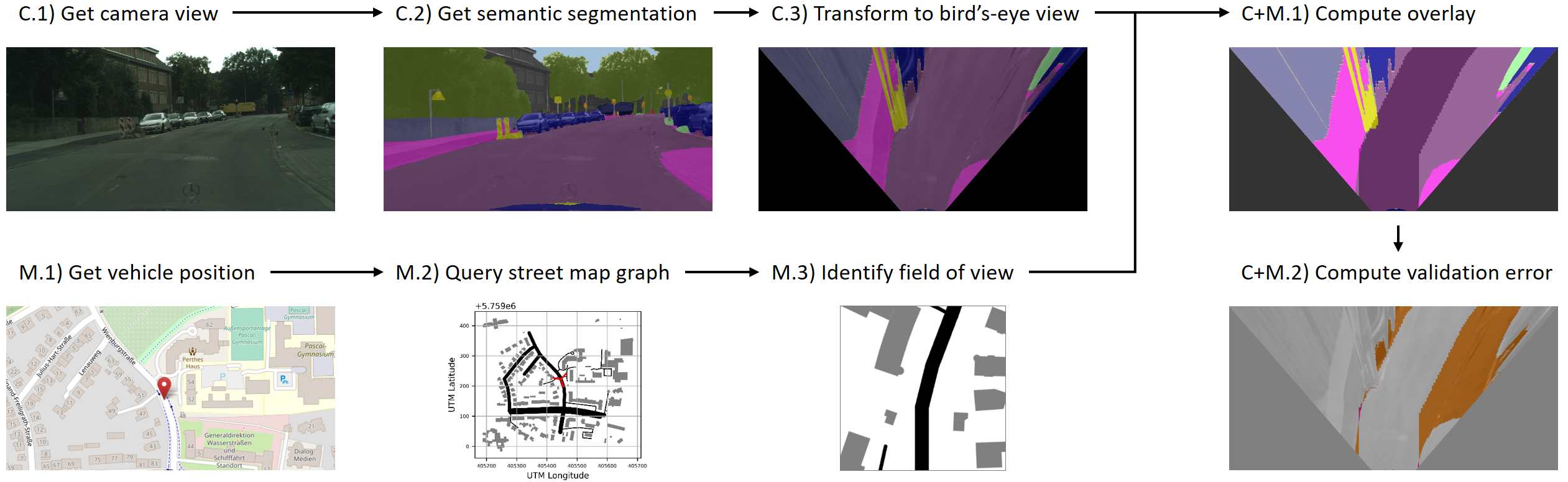}
\caption{
\textbf{Approach overview}: We validate the drivable area in semantic segmentation masks with a-priori knowledge from street maps. For this, we combine the segmentation mask of a given camera view with the street map that corresponds the given vehicle position. We compute an overlay of the segmentation in bird's-eye view and the street map's field of view, which we use to identify validation errors. The steps with \textit{C} describe tasks involving a \textit{c}amera view and the steps with \textit{M} describe tasks involving a \textit{m}ap view.
}
\label{fig:approach}
\end{figure*}

An advantage of our proposed method is that it facilitates a model-agnostic validation of deep learning approaches. It can be applied either offline in the testing phase of models, or even online to assess potentials errors within the current prediction, since the approach does not require ground truth to be available.
Another advantage is that it can be used to test models even in geographical regions that had not been represented in the training data.
This is relevant because the characteristics of semantic concepts such as roads, cars, or vegetation can be very diverse across different regions, but training datasets do not always reflect this domain variety~\cite{tsai2018learning, wang2020train}.
For autonomous driving, external data sources such as street maps can provide a valuable alternative information source for static objects present in ground truth data.

Our contributions in this paper are as follows. First, we present our approach to validate the drivable area in semantic segmentation masks using a-priori knowledge obtained from street maps.
Second, we show initial experimental results, which suggest a connection between segmentation errors and inconsistencies with the map, indicating that prediction errors can be identified not only with ground truth data but also with street maps.
Third, we discuss important challenges for this approach.
The paper is structured accordingly.

\section{Approach}

Our approach validates the drivable area in a given semantic segmentation mask using the corresponding geometric structures in a given street map. An overview of our approach is illustrated in Figure~\ref{fig:approach}. In the following we shortly describe each step within the approach.

\paragraph{Step C.1) Get camera view.} We retrieve an image from the vehicle's front view of a traffic scene. Here we use the Cityscapes~\cite{cordts2016cityscapes} dataset.

\paragraph{Step C.2) Get semantic segmentation.}
Using a neural network, we obtain a segmentation mask that maps each image pixel to a set of pre-defined class labels. In the example image in Figure~\ref{fig:approach}, the labels are visualized in a chosen color coding: \textit{road} is violet, \textit{pedestrian walk} is pink, \textit{car} is blue, etc. Here, we used a model trained by the ERFNet encoder-decoder architecture~\cite{romera2017erfnet} to create the predictions.

\paragraph{Step C.3) Transform to bird's-eye view.} To prepare the validation of the drivable area segments using the street map, we transform the segmentation image into a bird's-eye view, which corresponds to the view space of the street map. For this, we applied an image-based perspective transformation.

\paragraph{Step M.1) Get vehicle position.} We get the position by reading the GPS coordinates of the vehicle and thus retrieve latitude, longitude and the heading. These values are given in the Cityscapes dataset for each camera image. The accuracy of the GPS position can be a challenge, as discussed further in Section~\ref{sec:challenges}.

\paragraph{Step M.2) Query street map graph.} For the given latitude and longitude we get the street map graph for the surrounding area. For our analysis we use data from OpenStreetMap~\cite{OpenStreetMap}, because this source offers a freely available option with sufficient data coverage for initial experiments.
For future application in production other map providers might be preferable, as we discuss in Section~\ref{sec:challenges}.

\paragraph{Step M.3) Identify field of view.} We transform the street map graph to an image and rotate it in the direction of the vehicle's heading, which is obtained from the metadata of the camera image. We zoom in so that it corresponds to the potential field of view from the camera mounted on the car.

\paragraph{Step C+M.1) Compute overlay.} After both the camera view and the map segment are prepared, we create an overlay of the road from the map image with the semantic segmentation in bird's eye view. As shown in Figure~\ref{fig:approach}, the road is illustrated as a transparent black area. This allows us to recognize the overlap between the predicted road segments from the semantic segmentation mask and the street map.

\paragraph{Step C+M.2) Compute validation error.} Finally we compute the regions where the predictions of the trained model deviate from the a-priori knowledge contained in the map.
Two types of validation errors can be derived: False positive regions, i.e., where the segmentation shows a road, but the map does \textit{not} (here visualized by orange red), and false negative regions, i.e., where the segmentation does \textit{not} show a road, but the map does (visualized by pink red). For computing reliable error regions, we omit pixels that are assigned to labels that could be occluding the drivable area like, e.g., vegetation or cars.

\section{Results}

\begin{figure}[t]
    \centering
    \includegraphics[width=0.95\columnwidth]{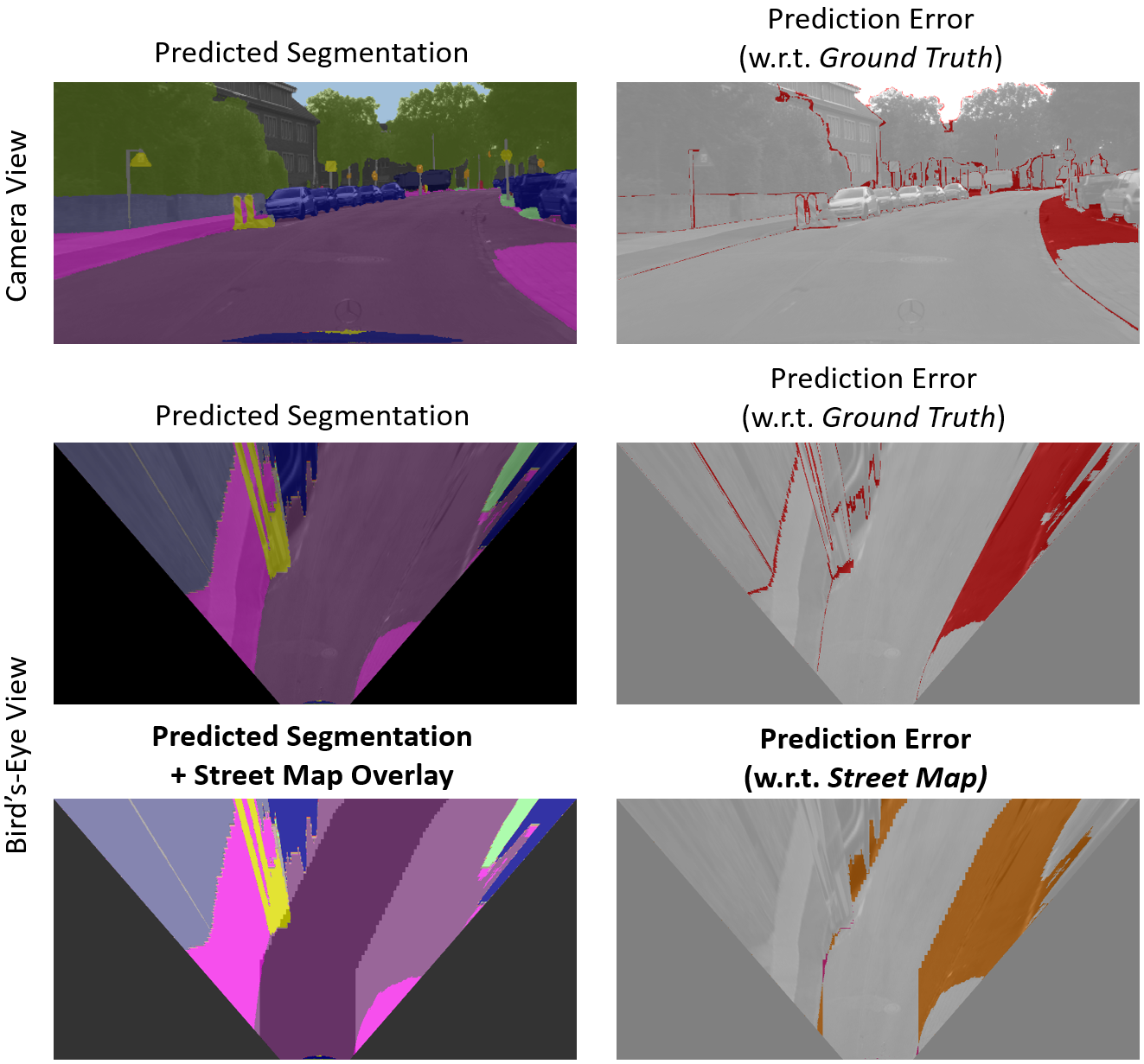}
    \caption{
    \textbf{Example 1 (False positive road).}
    This example shows a traffic scene for which the predicted segmentation shows a road that gets wider to the right, but according to the ground truth the road keeps its width and instead there is a pedestrian walk at the right side.
    This deviation results in a larger error region, which is highlighted in red color in the prediction error images (top and middle right).
    Our map-based validation approach also identifies this deviation:
    The street map suggests a road of constant width, resulting in a detected false positive road error region, which is highlighted in orange red color in the image of the error w.r.t. the street map (bottom right).
    }
    \label{fig:ex_fp1}
\end{figure}

In first experiments we see that a street map can indeed be used to validate drivable segments in segmentation masks. In our results we qualitatively observe that a map-based validation can identify similar error regions in the drivable area as an error evaluation of the prediction using ground truth data.

\begin{figure}[t]
    \centering
    \includegraphics[width=0.95\columnwidth]{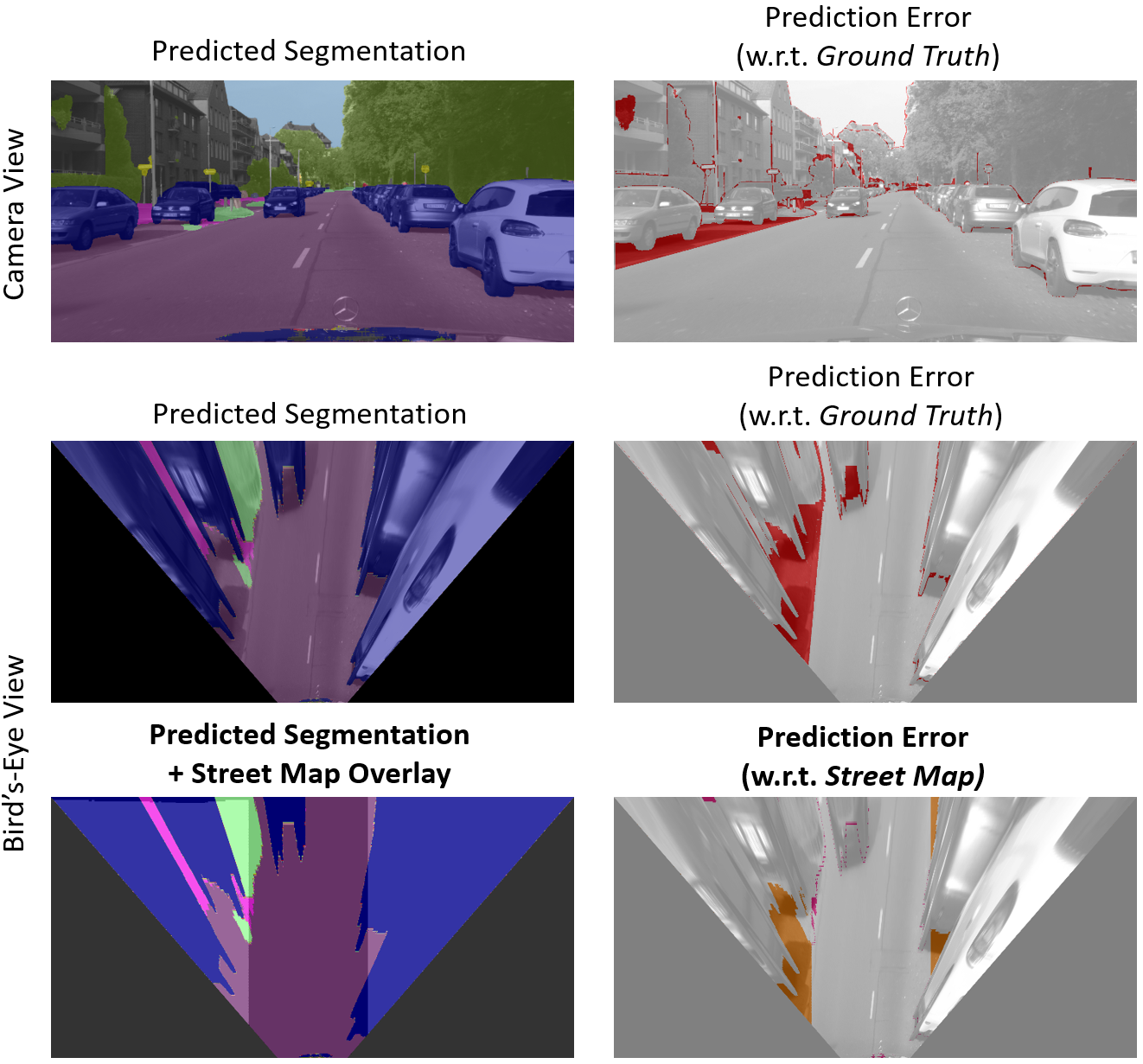}
    \caption{
    \textbf{Example 2 (False positive road).}
    The predicted segmentation shows a road straight forward and below the cars parked at the left side of the street. According to the ground truth there is a parking space below that parking cars.
    Our map-based validation approach identifies this deviation: The street map suggests a less broad road than in the prediction, resulting in a detected false positive region, which is highlighted in orange red color in the corresponding error image (bottom right).
    }
    \label{fig:ex_fp2}
\end{figure}

\begin{figure}[t]
    \centering
    \includegraphics[width=0.95\columnwidth]{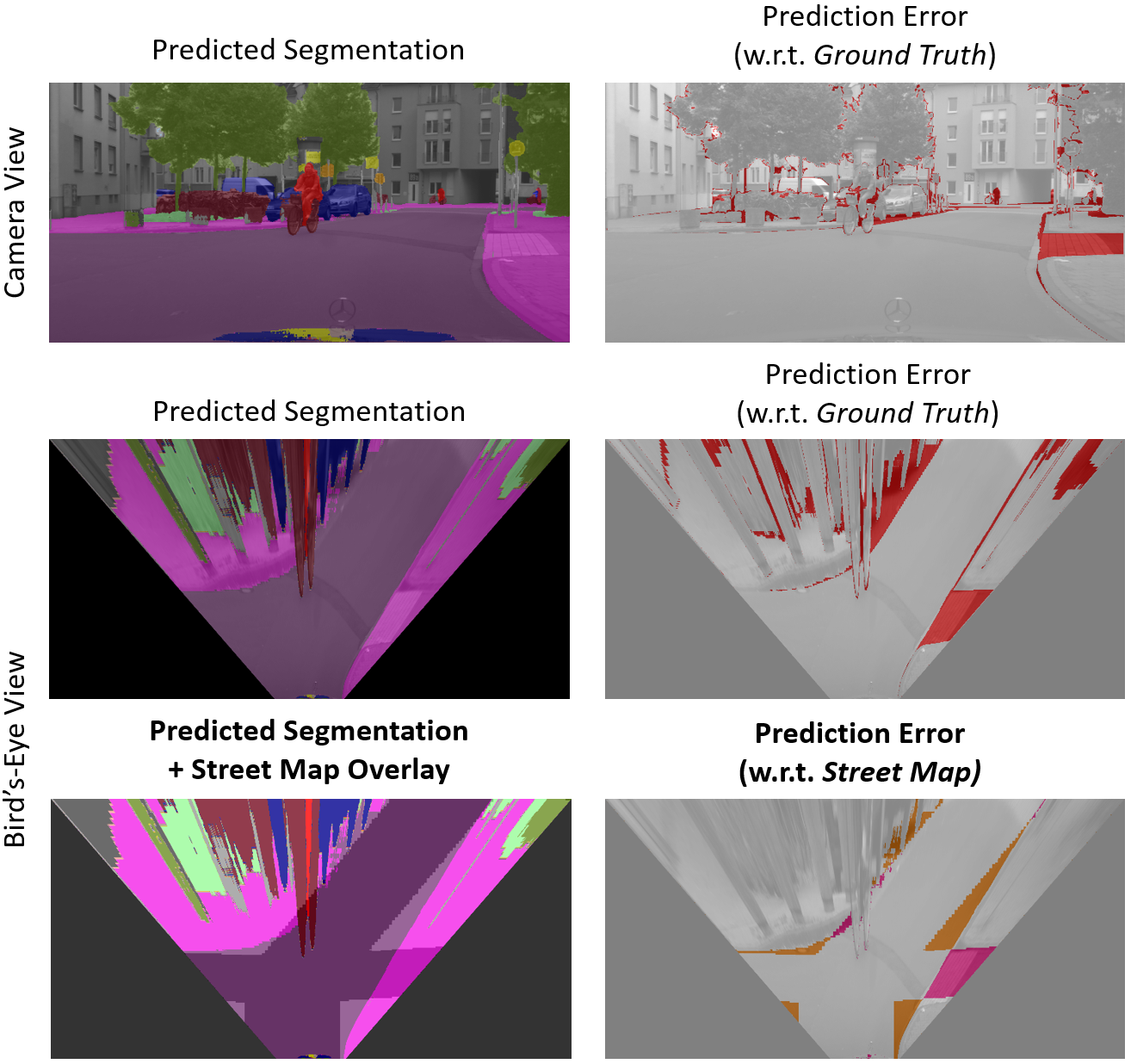}
    \caption{
    \textbf{Example 3 (False negative road).} The predicted segmentation shows a road running straight forward, although there is an intersection to the right according to the ground truth.
    Our approach identifies this deviation, too: The street map shows an intersection to the right. This results in a detected false negative region, which is highlighted in pink red color at the right side of the error image (bottom right).
    }
    \label{fig:ex_fn}
\end{figure}

Figures~\ref{fig:ex_fp1}, \ref{fig:ex_fp2}, and \ref{fig:ex_fn} show exemplary results for segmentations that contain errors in the prediction of the road segment.
The first rows show the predicted semantic segmentation (top left) and its error image with respect to the ground truth (top right).
The second rows show both images transformed to the perspective of the map, i.e., bird's-eye view, as a preparation for the combination with a street map.
The third rows show the overlay of the predicted segmentation with the corresponding street map (bottom left) and the error image of the predicted segmentation with respect to the street map (bottom right; false positive roads are orange red, false negative roads are pink red). Both error images, on the one side with respect to ground truth and on the other side with respect to the street map, highlight the falsely inferred regions.

While the examples in Figures~\ref{fig:ex_fp1} and~\ref{fig:ex_fp2} show a traffic scene where the segmentation predicts a road although there is no road (false positive), Figure~\ref{fig:ex_fn} shows a scene where a crossing road is not detected (false negative). In both cases our method can help to identify such errors.

Since first experimental results suggest that prediction errors with respect to the map are similar to those with respect to the ground truth,
it can be concluded that street maps could provide the possibility of validating segmentation masks via map data, which can be assumed to be available in many applications of segmentation networks.
This could be beneficial especially in situations where there is no ground truth available, e.g. for large amounts of unlabelled video sequences or potentially even as an online method.

\section{Challenges}
\label{sec:challenges}

The validation of segmentation masks using street map data poses challenges with respect to the localization precision and the map information density, which we briefly discuss next.

Inaccuracies in the vehicle localization, e.g. through a GPS position, would lead to inaccuracies in the correspondingly selected street map area.
However, modern and future technologies in autonomous driving cars like landmark detection and priors can provide a localization within a few centimeters~\cite{wilbers2019localization}.
In our experiments we employed the widely used Cityscapes~\cite{cordts2016cityscapes} dataset, for which we observed inaccuracies in the localization of GPS positioning.
Apart from this, the currently most used semantic segmentation datasets in relevant publications do not contain all the required vehicle localization data in terms of latitude, longitude and especially heading.
This brings up a challenge for data availability and quality.
For the first experiments, we decided to use the Cityscapes dataset due to its presence in current research and avoid GPS localization errors via a manual correction routine to the position, in order to demonstrate our general approach.
In future experiments and real-world applications, data with much more precise localization information should be used.

Furthermore, the information density of the map under consideration defines the share of the prediction that can be validated via map knowledge.
Although open map providers like OpenStreetMap integrate content from various contributors, which can easily be updated, such sources often lack details. These could include, for example,
the precise geometric curvatures or the exact width of a road, or
the presence of pedestrian walks next to it.
For the first experiments this map data seems to create decent results, however, we observed that missing information about street curvature can falsely increase the prediction error, as shown in Figure~\ref{fig:ex_fn}.
However, other map providers offer more detailed and accurate information and such high-definition maps should be used in future.


\section{Conclusion}

In this paper we proposed to validate machine learning models with a-priori domain knowledge and presented an approach that validates semantic segmentation masks with given street maps.
First experiments showed promising results and indicate that models can not only be tested with ground truth data, but also with independent prior knowledge leveraged from already available map data.


\section*{Acknowledgements}
Parts of this work have been supported by
the Research Center for Machine Learning (RCML) within the Fraunhofer Cluster of Excellence Cognitive Internet Technologies (CCIT)
and
the Competence Center for Machine Learning Rhine Ruhr (ML2R), which is funded by the Federal Ministry of Education and Research of Germany (grant no. 01|S18038A).
The team of authors gratefully acknowledges this support.




\footnotesize{
\bibliography{references}
\bibliographystyle{template/icml2020}
}





\end{document}